%% file: ms.tex
\documentclass{article}

\usepackage[preprint,nonatbib]{neurips_2020}

\input{mysnipets}

\usepackage[utf8]{inputenc} % allow utf-8 input
\usepackage[T1]{fontenc}    % use 8-bit T1 fonts
\usepackage{hyperref}       % hyperlinks
\usepackage{url}            % simple URL typesetting
\usepackage{booktabs}       % professional-quality tables
\usepackage{amsfonts}       % blackboard math symbols
\usepackage{nicefrac}       % compact symbols for 1/2, etc.
\usepackage{microtype}      % microtypography
\usepackage{graphicx}
\usepackage{multirow}
\usepackage{wrapfig}

\usepackage{cleveref}
\usepackage[%
frozencache=true,
cachedir=.,
newfloat=true,
]{minted}

\SetupFloatingEnvironment{listing}{name=Algorithm}
\Crefname{listing}{Algorithm}{Algorithms}
\usepackage[sortcites,
backend=biber,
hyperref=true,
style=numeric,
language=auto,
maxbibnames=20,
maxcitenames=2,
eprint=false]{biblatex}
\newcommand{\ourmethod}{Meta Approach to Data Augmentation Optimization\xspace}
\newcommand{\ourmethodshort}{MADAO\xspace}
\newcommand{\randa}{RandAugment\xspace}
\newcommand{\autoa}{AutoAugment\xspace}
\newcommand{\zero}{\hphantom{0}}

\title{\ourmethod}

\author{%
Ryuichiro Hataya$^{1,2}$, Jan Zdenek$^{1}$, Kazuki Yoshizoe$^{2}$, Hideki Nakayama$^{1}$ \\
$^{1}$ The University of Tokyo, $^{2}$ RIKEN AIP
}

\begin{document}

\maketitle

\begin{abstract}

Data augmentation policies drastically improve the performance of image recognition tasks, especially when the policies are optimized for the target data and tasks. In this paper, we propose to optimize image recognition models and data augmentation policies simultaneously to improve the performance using gradient descent. Unlike prior methods, our approach avoids using proxy tasks or reducing search space, and can directly improve the validation performance. Our method achieves efficient and scalable training by approximating the gradient of policies by implicit gradient with Neumann series approximation. We demonstrate that our approach can improve the performance of various image classification tasks, including ImageNet classification and fine-grained recognition, without using dataset-specific hyperparameter tuning.

\end{abstract}

\section{Introduction}

Data augmentation is an effective way to improve the performance of CNN models for image recognition tasks, particularly when its policy is optimized for the target dataset. Conventional data augmentation for images consists of image transformation operations, such as random cropping and flipping, and color enhancing including modification of color intensities \cite{Krizhevsky2012,He2016b}. However, designing good data augmentation strategies requires profound understanding of the target data and operations.
For example, CutOut \cite{DeVries2017} randomly erases a patch region of each image and is known to improve performance on the CIFAR-10 dataset, but is also reported to degrade the performance on other datasets, e.g., ImageNet \cite{Lopes2019}.

Therefore, automatically designing effective augmentation strategies according to target data and tasks is desirable to improve the performance of image recognition models. One approach to augment existing data is to generate new data samples by interpolating several training images \cite{Hauberg2015,Devries2017,Zhang2018} or by using conditional generative models \cite{Antoniou2018,Baluja2018}. However, this approach requires a large amount of labeled data \cite{Rather2017} and sometimes fails to improve the performance \cite{Ravuri2019}, even if powerful conditional generative models are used.
On the other hand, some methods improve the performance by efficiently selecting effective combinations of image transformation operations from exponentially large candidate pools \cite{Rather2017,Cubuk2018a,Zhang2020}. In particular, \autoa \cite{Cubuk2018a} and its family \cite{Ho2019,Lim2019,Hataya2019a,Lin2020,Cubuk2019,Berthelot2019a,Lin2020} optimized combinations of operations to improve validation performance and achieved state-of-the-art results. 

These methods involve a bi-level optimization: the inner process optimizes parameters of a CNN on training data using a given combination of operations, and the outer process optimizes the combination of operations to maximize the validation performance. Particularly, the inner loop, i.e., training of a CNN, is usually expensive. Therefore, prior works use proxy tasks that adopt small subsets of training datasets or small models \cite{Cubuk2018a,Ho2019,Lim2019,Hataya2019a} or reduce the search space \cite{Cubuk2019,Berthelot2019a,Lin2020} to keep the entire training feasible. 
Such approaches may result in sub-optimal solutions. We will further review and formalize this problem in \Cref{sec:generalizing_dao}.

In this paper, we tackle the original bi-level optimization problem directly without using proxy tasks or reducing the search space. We propose \ourmethod (\ourmethodshort), which optimizes CNNs and augmentation policies simultaneously by using gradient-based optimization. Here, policies are updated so that they directly increase CNNs' performance. 
Na\"ively applying gradient-based optimization to this bi-level optimization requires differentiation through the inner optimization process \cite{Finn2017b} or computation of the inverse Hessian matrix \cite{Bengio2000}, both of which suffer from large space complexity. These problems are fatal because data augmentation optimization needs to handle large networks, e.g., CNNs for ImageNet. We bypass these issues by using the implicit gradient method with Neumann series approximation. 
Thanks to these approximations, \ourmethodshort is simple with little overhead as shown in \Cref{ls:bilevel_optimization}. Notably, this simplicity allows \ourmethodshort to scale to problems of ImageNet size, which has been nearly impossible for existing bi-level optimization methods \cite{hutter19}.

We empirically demonstrate that \ourmethodshort learns effective data augmentation policies and achieves performance comparable or even superior to existing methods on benchmark datasets for image classification: CIFAR-10, CIFAR-100, SVHN and ImageNet, as well as four fine-grained datasets. In addition, we show that \ourmethodshort improves the performance of self-supervised learning on ImageNet. All of the reported results have been achieved \textit{without using dataset-specific configurations}.

\newcommand{\highlig}[1]{\colorbox[rgb]{1.0, 0.8, 0.8}{#1}}
\begin{listing}[t]
\centering

\begin{minted}[%
linenos,
xleftmargin=20pt,
breaklines,
frame=lines,
mathescape,
escapeinside=||,
]{python}
cnn.initialize() # parameterized by $\vtheta$
|\highlig{policy.initialize()}| # parameterized by $\vphi$
for epoch in range(num_epochs):
    for train_data, val_data in data_loader:
        for i in range(num_inner_iters):
            input = |\highlig{policy(}|train_data[i]|\highlig{)}| # augment data by policy
            criterion = cnn.train(input) # referred to $f$ in the text
            cnn.update(criterion)
        vcriterion = cnn.val(val_data) # referred to $g$ in the text
        |\highlig{policy.update(vcriterion)}|
\end{minted}
\vspace{-0.8\baselineskip}
\caption{Our proposed method \ourmethodshort in a Python-like pseudo code. \ourmethodshort optimizes \mintinline[]{python}{policy} as well as \mintinline[]{python}{cnn} by gradient descent with \textit{slight modifications} (\highlig{\texttt{highlighted}}) and \textit{little overhead} to the standard optimization protocol.}
\label{ls:bilevel_optimization}
\end{listing}

\section{Generalizing Data Augmentation Optimization} \label{sec:generalizing_dao}

In \Cref{sub:optimization_da}, we describe the preliminaries of data augmentation optimization of \autoa family, and then review prior methods by generalizing the problem in \Cref{sub:generalization}.

\subsection{Designing Data Augmentation Space}\label{sub:optimization_da}

\begin{figure}
    \centering
    \includegraphics[width=0.65\linewidth]{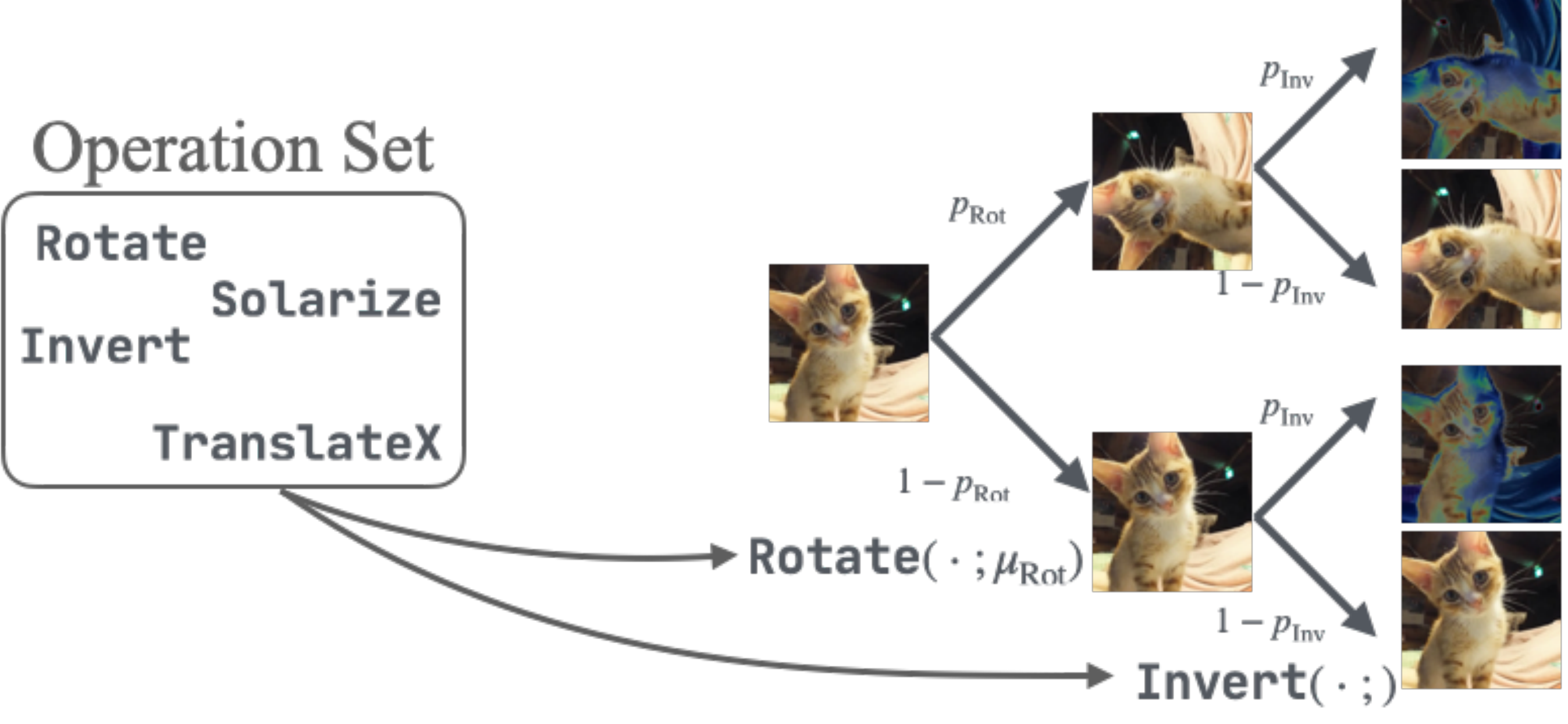}
    \vspace{-0.7\baselineskip}
    \caption{In \autoa family, a \mintinline[]{python}{policy} consists of operations. Each operation, e.g., $\texttt{Rotate}$ or $\texttt{Invert}$, augments an image with a probability of $p$ and a magnitude of $\mu$. As can been seen, performing multiple operations virtually increases the diversity of images. The operations are selected from given operation sets. Our proposed method \ourmethodshort learns to select suitable operations and their parameters, and simultaneously optimize a CNN by using gradient descent. This figure shows a 2-stage augmentation.}
    \label{fig:subpolicies}
\end{figure}

Let us define a set of input images $\gX$ and a set of operations $\gS$ consisting of data augmentation operations such as rotation and color inversion. In the \autoa family, each image $\vx\in\gX\subset[0, 1]^D$ is augmented by an operation $O: \gX\to\gX$ with a probability of $p_O\in[0, 1]$ and a magnitude of $\mu_O\in[0, 1]$ as illustrated in \Cref{fig:subpolicies}. The magnitude parameter can correspond for example to the degree of rotation, while some operations, such as inversion, have no magnitude parameter. By applying $K$ consecutive augmentation operations, each image results in $2^K$ possible images, that is, the number of images virtually increases. This formulation makes the size of the search space $(\lvert\gS\rvert\times[0, 1]\times[0, 1])^{K}$, where $\lvert\gS\rvert$ is the size of the operation set.

Operations and accompanied parameters need to be selected so that they minimize the validation criterion, such as the error rate. Usually, this selection is performed heuristically \cite{Krizhevsky2012,He2016b}. However, \citeauthor{Cubuk2018a} showed that data-driven optimization surpasses handcrafted selection \cite{Cubuk2018a}.

\subsection{Generalizing  AutoAugment Family}\label{sub:generalization}

Let $\vtheta\in\R^M$ denote parameters of a CNN model and $\vphi\in\R^N$ denote parameters of a policy for augmentation, i.e., selection of operations from the operation set and their accompanied parameters $\{(\mu_O, p_O); O\in\gS\}$. Let the empirical risk be $f(\vtheta, \vphi; \gD_T)$ and the validation criterion be $g(\vtheta; \gD_V)$. $\gD_T$ and $\gD_V$ are training and validation datasets. 
In the case of classification task, $\gD\coloneqq\{(\vx_i, y_i)\}\in\gX\times \{1, 2, \dots, C\}$, where $C$ is the number of classes.

Optimization of data augmentation policy in \autoa family methods can be generalized as

\vspace{-0.75\baselineskip}
\begin{equation}\label{eq:bi_level_optimization}
    \argmin_{\vphi}g(\argmin_{\vtheta}f(\vtheta, \vphi; \gD_T); \gD_V),
\end{equation}

\noindent that is, optimizing CNNs on training data with policies that minimize validation criteria on validation data.

Na\"ively solving this bi-level minimization problem takes a long time because CNN training $\min_{\vtheta}f(\vtheta, \vphi)$ is costly and the number of possible combinations of augmentation operations and their parameters is infeasibly large. Therefore, prior works tried to alleviate this problem in several ways. One direction is to reduce the search space over augmentation policies, i.e., dimension and range of the parameter $\vphi$ \cite{Cubuk2019,Berthelot2019a,Lin2020}. For example, \randa \cite{Cubuk2019} randomly samples operations from the operation set $\gS$ and shares $\mu_{O}$ among all operations. This reduction changes the outer problem in \Cref{eq:bi_level_optimization} from $\argmin_\vphi g$ to $\argmin_\mu g$, which makes it possible to use a simple searching process, such as grid search. OHL-AutoAug \cite{Lin2020} enables online searching using policy gradient \cite{Williams1992} by restricting the search space only to a limited range.

On the other hand, some methods use proxy tasks that approximate \Cref{eq:bi_level_optimization} to obtain (sub-) optimal policy $\vphi'$ to reduce the searching time of the inner optimization. The obtained policy $\vphi'$ is then used to train a CNN as $\min_{\vtheta}f(\vtheta, \vphi')$ \cite{Cubuk2018a,Ho2019,Lim2019,Hataya2019a} in an ``offline'' manner. 
For instance, \autoa employs a proxy task $f'$ that approximates the original inner problem $f$ as

\vspace{-0.75\baselineskip}
\begin{equation}\label{eq:proxy_task}
     \vphi'=\argmin_{\vphi}g(\argmin_{\vtheta'}f'(\vtheta', \vphi'; \gD'); \gD_V),
\end{equation}

\noindent with a smaller dataset $\lvert\gD'\rvert\ll\lvert\gD_T\rvert$ and a smaller network $\dim\vtheta'\ll\dim\vtheta$ for efficiency. The outer problem $g$ is optimized by black-box optimization techniques, such as reinforcement learning \cite{Cubuk2018a}. Fast \autoa \cite{Lim2019} and Faster \autoa \cite{Hataya2019a} approximate \Cref{eq:bi_level_optimization} as minimizing distance of distributions between augmented images and original images without directly minimizing $f$, which allows faster searching.

To summarize, prior methods indirectly solve the bi-level optimization problem in \Cref{eq:bi_level_optimization}, as displayed in \Cref{tab:ourmethod}. We instead propose to tackle this problem directly.

\begin{table}[t]
    \centering
    \caption{Comparison of prior methods and our proposal, \ourmethodshort. \textbf{\ourmethodshort can efficiently search \textit{full space} for policies on \textit{full datasets} with \textit{full CNN models}}, e.g., on ImageNet with ResNet-50.}
    \label{tab:ourmethod}
    \vspace{-0.5\baselineskip}
    \begin{tabular}{lcc}
       \toprule
           Method                            & Direct Inner Problem             & Direct Outer Problem \\  \midrule
           \autoa \cite{Cubuk2018a}          &                                  & \checkmark           \\
           Fast \autoa \cite{Lim2019}        &                                  & \checkmark           \\
           \midrule
           OHL-AutoAug \cite{Lin2020}        & \checkmark           & \\
           \randa \cite{Cubuk2019}           & \checkmark           & \\ 
           \midrule
           \ourmethodshort (ours)            & \checkmark           & \checkmark\\
       \bottomrule
    \end{tabular}
\end{table}

\section{Method} \label{sec:method}

In this paper, we propose to directly optimize the bi-level optimization problem in \Cref{eq:bi_level_optimization}. In other words, we optimize CNNs and augmentation policies \textit{simultaneously}, i.e., in an \textit{online} manner, without reducing the search space or using proxy tasks. With this simultaneous optimization, policies are expected to augment images according to the learning state of CNN models. \Cref{ls:bilevel_optimization} is a simple depiction of our approach, which we call \ourmethod, or in short, \ourmethodshort.

\subsection{Optimizing Policies by Gradient Descent}

\ourmethodshort directly optimizes the bi-level problem \Cref{eq:bi_level_optimization} using gradient descent. To this end, we assume that  $f$ and $g$ are differentiable \wrt $\vtheta$. Taking $C$-category classification as an example, cross entropy $-\E_{\vx_i, y_i}\log [\vh_{\vtheta}(\vx_i)]_{y_i}$ can be used as $f$ and $g$, but error rate $\E_{\vx_i, y_i}\1(\argmax \vh_{\vtheta}(\vx_i) \neq y_i)$ cannot. Here, $\1$ is the indicator function, and $\vh_{\vtheta}: \R^D\to\R^C$ is a CNN with a softmax output layer.

Gradient-based optimization of \Cref{eq:bi_level_optimization} requires $\nabla_{\vphi}g$ for iterative updating. Since the data augmentation implicitly affects the validation criterion, in other words, data augmentation is not used for validation, we obtain

\vspace{-0.6\baselineskip}
\begin{equation}\label{eq:grad_phi_g}
    \frac{\partial g}{\partial \vphi}=\frac{\partial g}{\partial \vtheta}\frac{\partial \vtheta}{\partial \vphi}.
\end{equation}

Because of the requirement of $g$, $\nabla_{\vtheta}g$ can be obtained. On the other hand, the exact computation of $\nabla_{\vphi}\vtheta$ has a large space complexity, as we will describe in \Cref{eq:unroll_steps}. Yet, if this gradient was available, the policies could be optimized by gradient descent.

\subsection{Approximating Gradients of Policy and Inverse Hessian}

To obtain $\nabla_{\vphi}\vtheta$ without suffering from a large space complexity, we can use the Implicit Function Theorem. If there exists a fixed point $(\vtheta^{\star}, \vphi^{\star})$ that satisfies $\nabla_{\vtheta}f(\vtheta^{\star}, \vphi^{\star})=0$, then there exists a function $\hat{\theta}$ around $\vphi^{\star}$ such that $\hat{\vtheta}(\vphi^\star)=\vtheta^{\star}$. If this condition holds, we also obtain

\vspace{-0.6\baselineskip}
\begin{equation}\label{eq:ift}
    \frac{\partial \hat{\vtheta}}{\partial \vphi} = -\left( \frac{\partial^2 f}{\partial \vtheta\partial \vtheta^{\top}} \right)^{-1} \frac{\partial^2 f}{\partial \vtheta\partial \vphi^{\top}}\Big|_{\vtheta^\star, \vphi^\star}.
\end{equation}

We can obtain an approximated gradient using this property. Unfortunately, $M=\dim\vtheta$ is usually large; therefore, computing the inverse Hessian matrix $(\nabla^2_{\vtheta}f)^{-1}$ is prohibitively expensive as it usually requires $\gO(M^3)$ computations. To avoid computing the inverse Hessian matrix, we use iterative methods based on the Neumann series, which boast better scalability than conjugate gradients in various problems \cite{Koh2017,Liao2018,Lorraine2019}.

The Neumann series $\mI+\mA+\mA^2+\dots=\sum_i^J \mA^j\to(\mI-\mA)^{-1} ~(J\to\infty)$ holds with a given squared matrix $\mA$ if $\|\mA\|<1$. Using this property, \Cref{eq:grad_phi_g} can be approximated with a positive integer $J$ as

\vspace{-0.75\baselineskip}
\begin{equation}
    \frac{\partial g}{\partial \vphi} = 
   -\frac{\partial g}{\partial \vtheta} 
    \left( \frac{\partial^2 f}{\partial \vtheta\partial \vtheta^{\top}}\right)^{-1} 
    \frac{\partial^2 f}{\partial \vtheta\partial \vphi^{\top}}
    \approx
   -\frac{\partial g}{\partial \vtheta}
    \sum_{j=0}^{J} \left(\mI-\frac{\partial^2 f}{\partial \vtheta\partial \vtheta^{\top}} \right)^j
    \frac{\partial^2 f}{\partial \vtheta\partial \vphi^{\top}}.
\end{equation}

We regularize the norm by simply introducing a scalar $\alpha\in\R^+$ as $\mI-\alpha\nabla_{\vtheta}^2 f$ as \cite{Lorraine2019}. We use $\alpha=10^{-3}$ with $J=5$ in the experiments. This Neumann series approximation can also help us avoid explicitly storing the Hessian matrix, whose space complexity is $\gO(M^2)$, by using Hessian-vector products derived from the following identity:

\vspace{-0.75\baselineskip}
\begin{equation}
    \vv^{\top}\left(\frac{\partial^2 f}{\partial \vtheta\partial \vtheta^{\top}}\right)
   =\frac{\partial}{\partial\vtheta}\left(\vv^{\top}\frac{\partial f}{\partial\vtheta}\right)\hphantom{xxxx}(\forall\vv\in\R^M).
\end{equation}

This right-hand side can be used instead of storing the Hessian matrix and only has the space complexity of $\gO(2M)$. 

\subsection{Differentiable Data Augmentation}\label{sub:dda}

To differentiate through $\vphi$, \ourmethodshort adopts the differentiable data augmentation pipeline following \cite{Hataya2019a}. As described in \Cref{sub:optimization_da}, each image $\vx$ is augmented with an operation $O$ with a magnitude of $\mu_O$ and a probability of $p_O$, which can be written as

\vspace{-0.75\baselineskip}
\begin{equation}\label{eq:operation_process}
    \vx \mapsto 
    \begin{cases}
        O(\vx; \mu_O)  &\text{ with probability of } p_O \\
        \vx            &\text{ with probability of } 1-p_O.
    \end{cases}
\end{equation}

This right-hand side can be written as $bO(\vx; \mu_O)+(1-b)\vx$ with a binary stochastic variable $b\sim\text{Bern}(b; p_O)$. Although the original Bernoulli distribution $\text{Bern}(b; p_O)$ is not differentiable \wrt $p_O$, Gumbel trick \cite{Jang2016} relaxes this restriction to enable backpropagation to update $p_O$. Similarly, some color-enhancing operations are non-differentiable \wrt the magnitude $\mu_O$, so the straight-through estimator \cite{Bengio2013} is used for such operations. We clamp $\mu_O$ and $p_O$ by a sigmoid function to limit their range to [0, 1].

% A notable difference between Faster \autoa and \ourmethodshort is the way of selecting operations. 
\ourmethodshort uses a different method to select operations compared to Faster \autoa in order to accelerate training.
\ourmethodshort selects operations using categorical distribution parameterized by a weight parameter $\vpi\in[0, 1]^{\lvert\gS\rvert}$, where $\sum_i\pi_i=1$. Since the original categorical distribution is non-differentiable as Bernoulli distribution, we use Gumbel-softmax with a temperature of $\tau\in\R^{+}$. This distribution, referred to as $\text{RelaxCat}(\vpi; \tau)$, samples one-hot-like vectors as $\tau\longrightarrow 0$. Using this distribution, an operation is selected and applied as

\vspace{-1.2\baselineskip}
\begin{align}
    \vx &\mapsto \frac{u_i}{\text{SG}(u_i)}O_i(\vx; \mu_{O_i}, p_{O_i}), \label{eq:transformation} \\
    i &= \argmax\vu, \\
    \vu &\sim \text{RelaxCat}(\vpi; \tau).
\end{align}

Here, $\text{SG}$ is the stop gradient operation, and thus, $\frac{u_i}{\mathrm{SG}(u_i)}=1$ so that the transformation \Cref{eq:transformation} keeps the range in $(0, 1)$. %$O_i$ is the selected operation. 
Different from this approach, Faster \autoa applies all operations and takes the weighted sum of the outputs to approximate this selection.

\subsection{Connection to Gradient-based Hyperparameter Optimization}

As can been seen, \Cref{eq:bi_level_optimization} is a hyperparameter optimization (HO) problem to neural networks. Traditional HO methods, such as grid search, random search \cite{Bergstra2012} and Bayesian optimization \cite{Snoek2012}, have poor scalability to increasing dimensionality of hyperparameters \cite{hutter19}. For that reason, gradient-based HO attracts attention. From HO viewpoint, policy parameters $\vphi$ are hyperparameters.

As shown in \Cref{eq:grad_phi_g}, we need to obtain $\nabla_{\vphi}\vtheta$ to optimize the outer problem $g$.  The inner optimization process (\Cref{ls:bilevel_optimization} Line 5-8) can be rewritten as 

\vspace{-0.75\baselineskip}
\begin{equation}\label{eq:unrolling}
\vtheta_T
=\vtheta_{T-1}-\eta\nabla_{\vtheta} f(\vtheta_{T-1}, \vphi)
=\cdots=
\vtheta_0-\eta\nabla_{\vtheta}\sum_{t=0}^{T-1} f(\vtheta_{t}, \vphi)
\eqqcolon \vtheta_T(\vtheta_0, \vphi)
\end{equation}

\noindent after $T$ SGD steps with learning rate of $\eta$. One approach to obtain $\nabla_{\vphi}\vtheta$ is to unroll the steps in \Cref{eq:unrolling} as  \cite{Maclaurin2015,Franceschi2018,Shaban2018}:

\vspace{-0.75\baselineskip}
\begin{equation}\label{eq:unroll_steps}
    \nabla_{\vphi}\vtheta\approx\nabla_{\vphi}\vtheta_T(\vtheta_0, \vphi).
\end{equation}

This unrolling requires to cache $\vtheta_0, \dots, \vtheta_{T-1}$, and thus, the space complexity becomes $\gO(TM)$, which might be prohibitive for large neural networks, while \ourmethodshort requires $\gO(2M)$.

Alternatively, implicit gradient yields $\nabla_{\vphi}\vtheta$ as explained in \Cref{eq:ift}. This approach needs the inverse Hessian matrix $(\nabla^2_{\vtheta}f)^{-1}$ \cite{Bengio2000}, but this computation is infeasible for modern neural networks with millions of parameters. Iterative methods, such as conjugate gradient \cite{Do2009,Domke2012,Pedregosa2016a,Rajeswaran} or Neumann series approximation \cite{Lorraine2019}, effectively compensate for this issue by approximating this inverse matrix in gradient hyperparameter optimization. Such iterative approximation methods using the Neumann series are also used in approximating influence function \cite{Koh2017} and enabling RNNs to handle long sequences \cite{Liao2018}. We exploit the knowledge from these prior works and adopt the implicit gradient method with Neumann series approximation to efficiently handle large-scale datasets and CNNs.

\section{Experiments and Results} \label{sec:experiments}

This section describes the empirical results of our proposed method in supervised learning for image classification and self-supervised learning tasks. For classification tasks, we use CIFAR-10, CIFAR-100 \cite{Krizhevsky2009}, SVHN \cite{Netzer2011} and ImageNet (ILSVRC-2012) \cite{Russakovsky2015}. In addition, we also used fine-grained classification datasets: Oxford 102 Flowers \cite{Nilsback08}, Oxford-IITT Pets \cite{parkhi12a}, FGVC Aircraft \cite{maji13} and Stanford Cars \cite{Krause2013}. For the self-supervised learning task, we used ImageNet. In all experiments except those on ImageNet, we set 10 \% of the original training data aside as validation data $\gD_V$ and report error rates on test data. For ImageNet, we used 1 \% of the training data as validation data. Note that this data split means that we use less training data than prior works and that changes the performance of baseline models. More details about experiment configurations are in \Cref{ap:exp_details}.

We used 14 operations for augmentation: \texttt{ShearX}, \texttt{ShearY}, \texttt{TranslateX}, \texttt{TranslateY}, \texttt{Rotate}, \texttt{Invert}, \texttt{AutoContrast}, \texttt{Equalize}, \texttt{Solarize}, \texttt{Color}, \texttt{Posterize}, \texttt{Contrast}, \texttt{Brightness} and \texttt{Sharpness} (see \Cref{ap:operations} for more details.) To make these operations differentiable, we implemented them using PyTorch \cite{NIPS2019_9015} and kornia \cite{eriba2019kornia}. We scale the magnitude of operations from $0$ to $1$ so that the magnitude of $0$ means no change and the magnitude of $1$ corresponds to the strongest level of the particular augmentation operation. We initialized the parameters of magnitude and probability with $\text{sigmoid}(0.5)\approx0.62$ and the parameters for operation selection to be equal. That means that MADAO in its initial state is nearly equivalent to \randa with the magnitude of $0.62$. We set the number of augmentation stages $K=2$ in all experiments below (see \Cref{fig:subpolicies}).

For training, we used SGD with a momentum for CNN optimization and RMSprop optimizer with learning rate of $10^{-2}$ for policy optimization, following \cite{Lorraine2019}. \Cref{eq:ift} requires the existence of fixed points that satisfies $\nabla_{\vtheta}f(\vtheta, \vphi)=0$. Following \cite{Lorraine2019}, we assume that $s$ iterations, which corresponds to \mintinline[]{python}{num_inner_iters} in \Cref{ls:bilevel_optimization}, makes the parameters hold the condition. To further encourage parameters to satisfy the condition, we also used warm up, i.e., training CNNs without augmentation for the first $w$ epochs. We set $s=30$ and $w=20$ by default. Validation criterion $g$ is the same loss function that is used as the training loss function $f$, e.g., cross-entropy, for simplicity.

Training of WideResNet 28-2 on CIFAR-10 for 200 epochs with \ourmethodshort takes 1.7 hours, while training without \ourmethodshort takes 1.0 hours in our environment. As regards memory consumption, training with \ourmethodshort requires 3.38 GB of GPU memory, while training without \ourmethodshort requires 1.74 GB in this setting. We tune the magnitude parameter of \randa using three runs of random search and report the test error rate for the run with the lowest validation error rate. 

% \subsection*{CIFAR-10, CIFAR-100, SVHN and ImageNet}
\subsection*{CIFAR-10, CIFAR-100 and SVHN}

\Cref{tab:results_small} presents test error rates on CIFAR-10, CIFAR-100 and SVHN with various CNN models: WideResNet-28-2, WideResNet-40-2, WideResNet-28-10 \cite{Zagoruyko} and ResNet-18 \cite{He2016b}. We show the average scores of three runs. For comparison, we present the results of \randa and the default augmentation: Cutout \cite{DeVries2017}, random cropping, and random horizontal flipping (except SVHN), following \cite{Cubuk2019}, which we refer to as Baseline. For CIFAR-10 and CIFAR-100, we also present the results of \autoa, which are trained on the split training set using the policy provided by the authors \footnote{\url{https://github.com/tensorflow/models/tree/master/research/autoaugment}}. \randa and \autoa are selected here as representative methods that limit the search space and use proxy tasks, as discussed in \cref{sub:generalization}.

As can been seen, \ourmethodshort achieves performance superior to baseline methods, which demonstrates the effectiveness of directly solving the bi-level problem rather than using proxy tasks as \autoa or a reduced search space as \randa.

\subsection*{ImageNet}

\Cref{tab:results_imagenet} shows the results on ImageNet with ResNet-50 \cite{He2016b} trained in supervised and self-supervised learning manners. In the supervised learning setting, we trained models for 180 epochs. In addition, to showcase the effectiveness of \ourmethodshort in other tasks than supervised learning, we apply \ourmethodshort to contrastive self-supervised learning \cite{he2019moco} for 100 epochs of training. We report the results of the linear classification protocol.

Most importantly, \ourmethodshort can scale to an ImageNet-size problem without using proxy tasks or reducing search space. We hypothesize that the slightly inferior performance of \ourmethodshort compared to \randa could be attributed to the fact that ImageNet has 1,000 categories and a single policy might be insufficient for such diverse data. We also believe that class- or instance-conditional data augmentation might be required to further improve performance, which we leave as an open problem.

\subsection*{Fine-grained Classification}

To showcase the ability of \ourmethodshort to augment images according to target datasets, we conducted experiments on four fine-grained datasets with ResNet-18 \cite{He2016b}. \Cref{tab:results_finedgrained} shows the average test error rates over three runs. These datasets are nearly ten to twenty times smaller than CIFARs and SVHN, yet \ourmethodshort drastically improves the performance without using dataset-specific hyperparameters. 
This improvement emphasizes the importance of searching for good policies according to full target datasets from the full search space. \ourmethodshort can capture the characteristics of each fine-grained dataset and generate tailored policies for each dataset from a large search space.

\begin{table}[tb]
    \centering
    \caption{Test error rates on CIFAR-10, CIFAR-100 and SVHN. For AutoAugment${}^{\ast}$, we train CNN models with the official \autoa policy for CIFAR-10 provided by the authors. \textbf{\ourmethodshort uses the same hyperparameters for all configurations and achieves superior performance to baselines, which demonstrates the effectiveness of our method.}}
    \vspace{-0.5\baselineskip}
    \label{tab:results_small}
    \begin{tabular}{llcccc}
       \toprule
       Dataset   &  Model            & Baseline  & AutoAugment${}^{\ast}$   & \randa      & \ourmethodshort (ours) \\ 
       \cmidrule(r){1-2} \cmidrule(rl){3-6}
       CIFAR-10  &  WideResNet 28-2  & 4.8       &    4.7              & 4.7          &  4.6          \\
                 &  WideResNet 40-2  & 4.5       &    4.3              & 4.3          &  4.2          \\
                 &  WideResNet 28-10 & 3.4       &    3.1              & 3.1          &  3.1          \\
                 &  ResNet-18        & 5.2       &    4.5              & 4.5          &  4.4          \\
       \cmidrule(r){1-2} \cmidrule(rl){3-6}
       CIFAR-100 &  WideResNet 28-2  & 24.8      &   24.0              & 24.0         &  23.9           \\
       \cmidrule(r){1-2} \cmidrule(rl){3-6}               
       SVHN      &  WideResNet 28-2  & 2.9       &    N/A              & 2.6           & 2.5             \\
       \bottomrule                                        

    \end{tabular}
\end{table}

\begin{table}[tb]
    \centering
    \caption{ImageNet test error rates (Top-1/Top-5) on supervised and self-supervised learning.}
    \vspace{-0.5\baselineskip}
    \label{tab:results_imagenet}
    \begin{tabular}{llccc}
       \toprule
       Task                 &  Model            & Standard & \randa       & \ourmethodshort (ours) \\ 
       \cmidrule(r){1-2} \cmidrule(rl){3-5}
       Supervised Learning       &  ResNet-50        & 23.7/ 6.9  &  22.5/ 6.4    & 23.0/ 6.6             \\
       \midrule
       Self-Supervised Learning  &  ResNet-50        & 42.1/19.4 &  39.6/17.6   & 39.6/17.7            \\
       \bottomrule
    \end{tabular}
\end{table}

\begin{table}[tb]
    \centering
    \caption{Test error rates on fine-grained datasets. \ourmethodshort outperforms \randa with a large margin. \textbf{Note that we use the same hyperparemeters as in the experiments in \Cref{tab:results_small}.}}
    \vspace{-0.5\baselineskip}
    \label{tab:results_finedgrained}
    \begin{tabular}{ccccccc}
       \toprule
       Dataset   &  Model            & \# Classes  & \# Training Set & Standard & \randa       & \ourmethodshort (ours) \\ 
       \cmidrule(r){1-4} \cmidrule(l){5-7}
       Flower    &  ResNet-18        & 102         &  2,040          & 10.7      &  \zero8.5   &  \zero7.5             \\
       Pet       &  ResNet-18        & \zero37     &  3,680          & 15.0      &  13.4       &  12.5                 \\
       Aircraft  &  ResNet-18        & \zero70     &  6,667          & 12.8      &  \zero9.7   &  \zero8.7            \\
       Car       &  ResNet-18        & 196         &  8,144          & 14.2      &  11.3       &  10.8                 \\
       \bottomrule
    \end{tabular}

\end{table}

\section{Analysis}

\subsection{How Policies Develop during Training}

\begin{figure}[tb]
    \centering
    \includegraphics[width=0.9\linewidth]{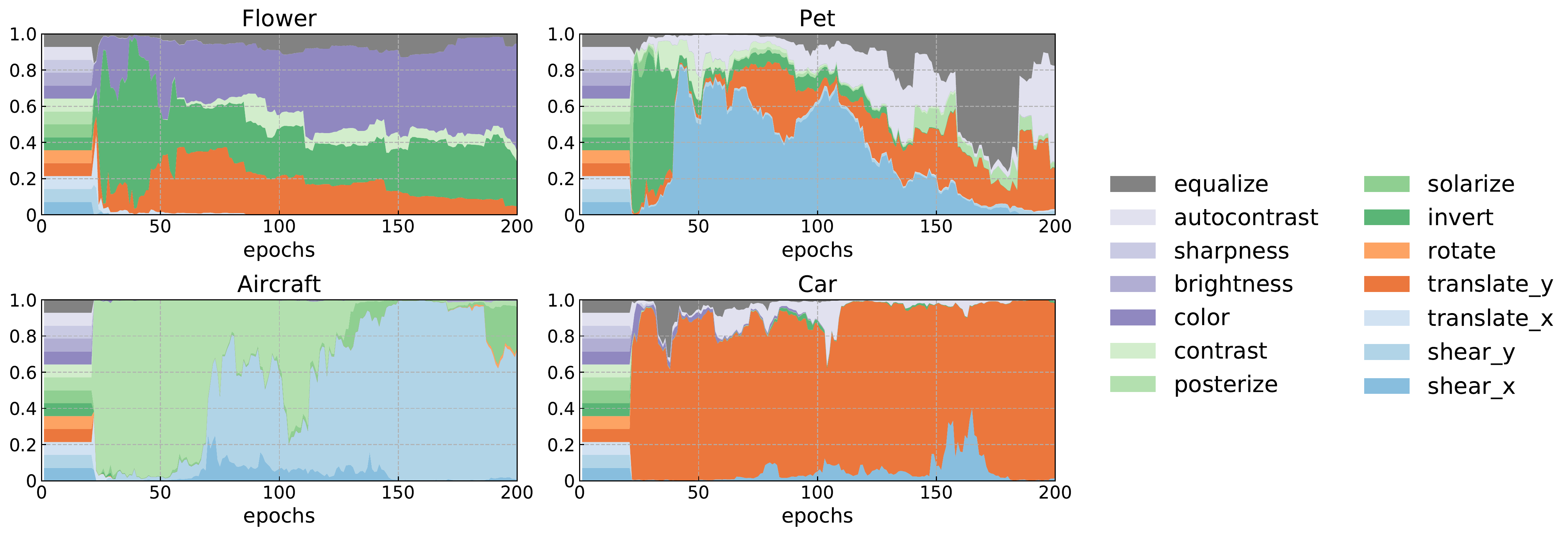}
    \vspace{-1\baselineskip}
    \caption{The development of operation selection probabilities at the first augmentation stage during training on fine-grained datasets for 200 epochs. Note that we set the first 20 epochs as a warm-up period during which parameters are not updated.}
    \label{fig:policy_development}
\end{figure}

In \Cref{fig:policy_development}, we present the development of policies during training on fine-grained datasets. As can be seen, each dataset has its specific operations that are selected, which could be thought as reflecting the characteristics of each dataset. Besides, the way of selection changes according to the learning phase. Note that the first 20 epochs are set to warm-up and the augmentation policy parameters are not updated. We show further observations in \Cref{ap:additional_experiments}.

\begin{wrapfigure}[11]{r}{0.4\linewidth}
    \centering
    \vspace{-3.5\baselineskip}
    \includegraphics[width=\linewidth]{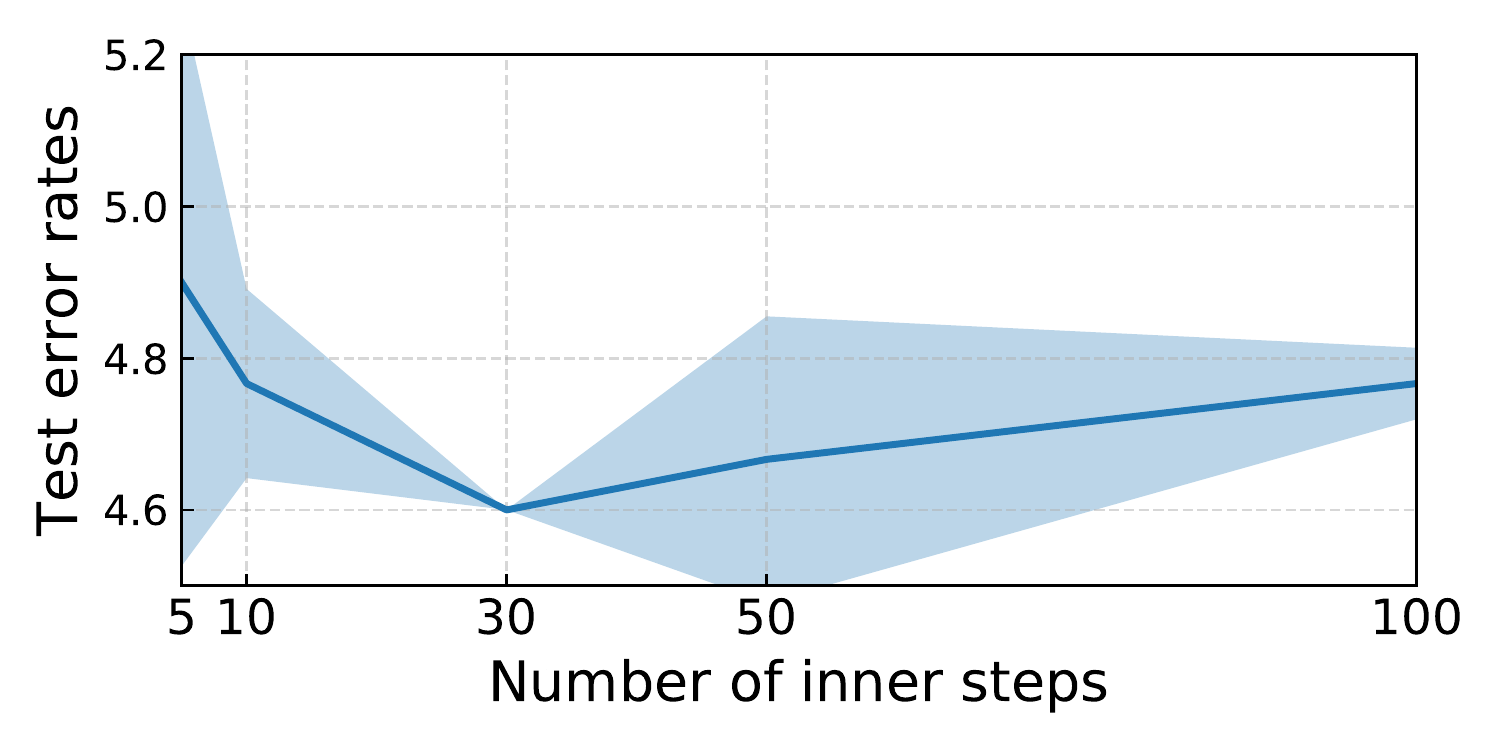}
    \vspace{-2\baselineskip}
    \caption{The relationship between the number of inner steps and the final test error rates on CIFAR-10 with WideResNet-28-2. We present the mean and standard deviation over three runs.}
    \label{fig:ablations}
\end{wrapfigure}

\subsection{How Inner Steps Affect the Performance}\label{sub:num_inner_optim}

\Cref{fig:ablations} presents test error rates with various numbers of inner update steps $s$ using WideResNet-28-2 on CIFAR-10. CNNs yield the best performance when $s=30$ and the results indicate that there exists a trade-off between ``exploration and exploitation'' of obtained policies: a small number of inner steps might not correctly evaluate the current policies, while running a large number of inner steps might fail to explore better strategies. Importantly, unrolled-based implementations would require to store $s$ model caches, which is infeasible for $s=30$ with modern CNNs. On the other hand, \ourmethodshort can efficiently handle a large $s$.

\section{Conclusion}

In this paper, we have proposed \ourmethodshort, a novel approach to optimize an image recognition model and its data augmentation policy simultaneously. To efficiently achieve this goal, we use the implicit gradient method with Neumann series approximation. As a result, the overhead of \ourmethodshort to the standard CNN training, w.r.t., time and memory, is marginal, which enables ImageNet-scale training. Empirically, we demonstrate on various tasks that \ourmethodshort achieves superior performance to prior works without restricting search space or using sub-optimal proxy tasks.

Data augmentation is known to boost the performance in various visual representation learning settings, such as semi-supervised learning \cite{Berthelot2019a,Suzuki2020}, domain generalization \cite{Volpi2018}, and self-supervised learning \cite{Chen2020}. We believe that our method can be introduced into these representation learning methods and efficiently enhance their performance, as we showcased with self-supervised learning in this paper.

\clearpage
\makeatletter
\if@preprint
    % nothing
\else
\section*{Potential Impacts of Our Work}

\ourmethodshort boosts performance of image recognition models with little overhead and minimal hyperparameter tuning. These advantages help to reduce energy consumption, which is a known issue of large-scale hyperparameter optimization, such as neural architecture search \cite{strubell2019}. \ourmethodshort showcases superior performance, especially on fine-grained datasets. This property enables further application of image recognition models to various domains, but it may also help in misuse of image recognition.

\fi
\makeatother

% bibliography
\printbibliography

\appendix

% \noindent Supplemental materials for \ourmethod.

\section{Operations Used in Policies}\label{ap:operations}

We introduce the operations used in \ourmethodshort in \Cref{ap:tab:operations}. Internally, magnitudes are restricted within $(0, 1)$ range with sigmoid function and rescaled to the appropriate range. For example, we multiply the internal magnitude $\mu_{\texttt{ShearX}}$ for \texttt{ShearX} operation by $0.3$. As can been seen, there are three operations that have no magnitude parameters. Therefore, each policy has $(11+14+14)\times K$ learnable parameters, where 11 corresponds to the number of magnitude parameters, e.g., $\mu_{\texttt{ShearX}}$, the first 14 corresponds to the number of probability parameters, e.g., $p_{\texttt{ShearX}}$, and the second 14 corresponds to operation selection parameters $\vpi$. In our experiments, we set $K=2$, and thus, the total number of learnable parameters is 78. Note that the original implementation of \randa does not include \texttt{Invert} in the operation set but we perform experiments with \randa using the same operation set as we use for our proposed method, that is including \texttt{Invert}.

\begin{table}[h]
    \centering
    \caption{Operations used in \ourmethodshort.}
    \label{ap:tab:operations}
    \begin{tabular}{llc}
    \toprule
               & Operation & Original Magnitude Range \\ \midrule
               
       \multirow{5}{*}{\shortstack[l]{Affine\\Transformation}}
       
             & \texttt{ShearX}       & $[0, 0.3]$                     \\
             & \texttt{ShearY}       & $[0, 0.3]$                     \\
             & \texttt{TranslateX}   & $[0, 0.45]$                    \\
             & \texttt{TranslateY}   & $[0, 0.45]$                    \\
             & \texttt{Rotate}         & $[0, 30]$                      \\ 
             \midrule
            
       \multirow{9}{*}{\shortstack[l]{Color\\Enhancing\\Operations}}
       
             & \texttt{Invert} & none                                   \\
             & \texttt{AutoContrast} & none                           \\
             & \texttt{Equalize}       & none                           \\
             & \texttt{Solarize}       & $[0, 256]$                     \\
             & \texttt{Color}          & $[0, 2]$                       \\
             & \texttt{Posterize}      & $[0, 4]$                       \\
             & \texttt{Contrast}       & $[0, 2]$                       \\
             & \texttt{Brightness}     & $[0, 2]$                       \\
             & \texttt{Sharpness}      & $[0, 2]$                           \\
    
    \bottomrule
    \end{tabular}
\end{table}

\section{Experimental Details}\label{ap:exp_details}

\begin{table}[h]
    \centering
    \caption{Shared hyperparameters in the experiments in \Cref{sec:experiments}.}
    \label{ap:tab:hyper_parameters}
    \begin{tabular}{ccc}
    \toprule
    Name               & Description                       & Shared Value                        \\
    \midrule
    Number of inner steps        & $s$ corresponds to \texttt{num\_inner\_iters} in \Cref{ls:bilevel_optimization}   &     30     \\
    Warm-up epochs      & Initial $w$th epochs that policy is not updated     &     20     \\
    Temperature        & $\tau$ for operation selection                      &     0.05   \\
    \bottomrule
    \end{tabular}
\end{table}

\subsection{CIFAR-10, CIFAR-100 and SVHN}

On CIFAR-10 and CIFAR-100, we trained WideResNets and ResNet-18 for 200 epochs. We used SGD with the initial learning rate of 0.1, the momentum of 0.9 and the weight decay of $5\times10^{-4}$. The learning rates ware scheduled with cosine annealing with warm restart \cite{Loshchilov2016}. On SVHN, we trained WideResNet 28-2 for 160 epochs. We used SGD with the initial learning rate of $5\times10^{-3}$, the momentum of 0.9 and the weight decay of $1\times10^{-4}$. The learning rate is divided by 10 at 80th and 120th epochs. On CIFAR-10, CIFAR-100 and SVHN, we set the batch size to 128.

\subsection{ImageNet}

\subsubsection*{Supervised Learning}
On ImageNet, we trained ResNet-50 for 180 epochs with SGD of the base initial learning rate of 0.1, the momentum of 0.9 and the weight decay of $1\times10^{-4}$. The learning rate is divided by 10 at 60th, 120th and 160th epochs. We set the batch size to 1,024 so that we scale the initial learning rate to 0.4. As the standard data augmentation, we randomly cropped images into $224\times224$ pixels and randomly flipped horizontally.

\subsubsection*{Self-supervised Learning}
We follow the experimental settings of MoCo \cite{he2019moco} and trained for 100 epochs with the batch size of 512 and the queue size of 65,536. We applied augmentation to images for both key and query networks. For linear classification, we used SGD with the momentum of 0.9 and set the initial learning rate to 30, which is decayed by 10 at 60th and 80th epoch.

\subsection{Fine-grained classification}

On fine-grained datasets, we trained ResNet-18 for 200 epochs and set the batch size to 64. As the standard data augmentation, we used the same strategy to ImageNet, including random cropping into $224\times224$ pixels.

\section{Additional Results}\label{ap:additional_experiments}

%Here, we show some additional experimental results. 

\subsection{How Policies Develop during Training}

\begin{figure}[t]
    \centering
    \includegraphics[width=\linewidth]{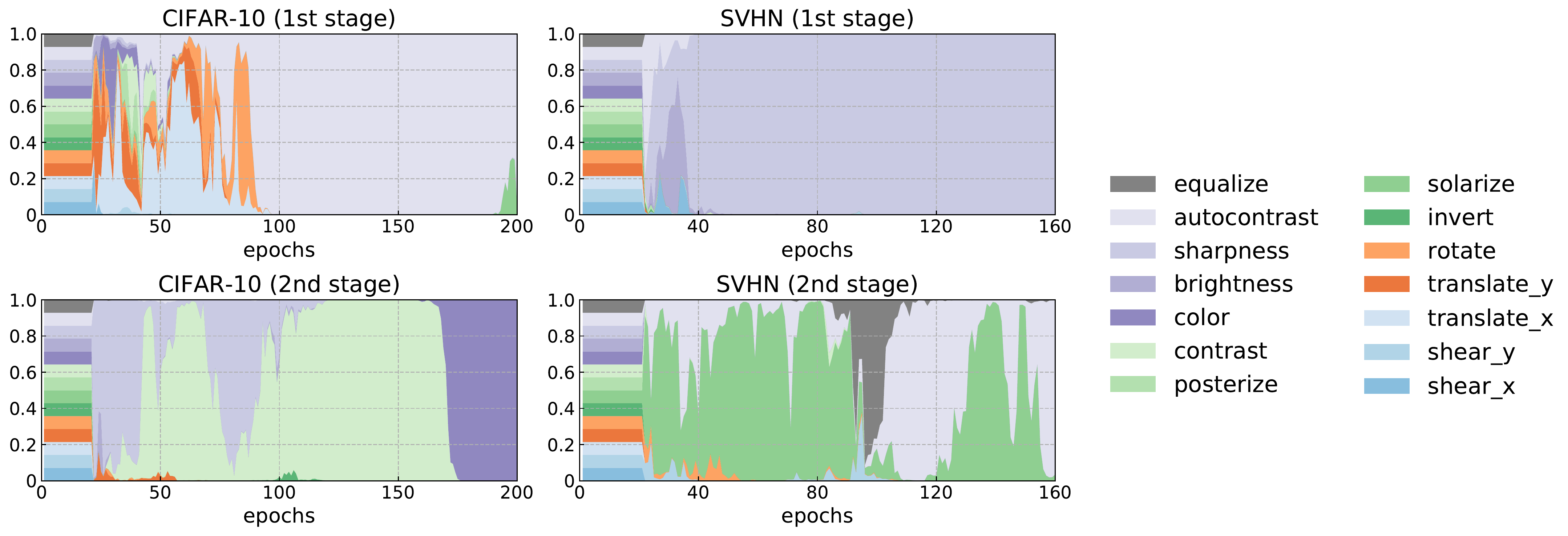}
    \caption{The development of the selection probabilities for each operation during training on CIFAR-10 and SVHN with WideResNet-28-2 for 200 epochs. Note that we set the first 20 epochs to warm-up period that parameters are not updated.}
    \label{ap:fig:policy_development}
\end{figure}

\Cref{ap:fig:policy_development} shows how the selection probabilities for each operation develop during training on CIFAR-10 and SVHN. Similar to fine-grained datasets shown in \Cref{fig:policy_development}, the policies for CIFAR-10 and SVHN also show clear difference to each other. As can be observed,  the first and second stage for each dataset evolve differently, which indicates that the stages develop complementarily to each other.

\begin{figure}[t]
    \centering
    \includegraphics[width=\linewidth]{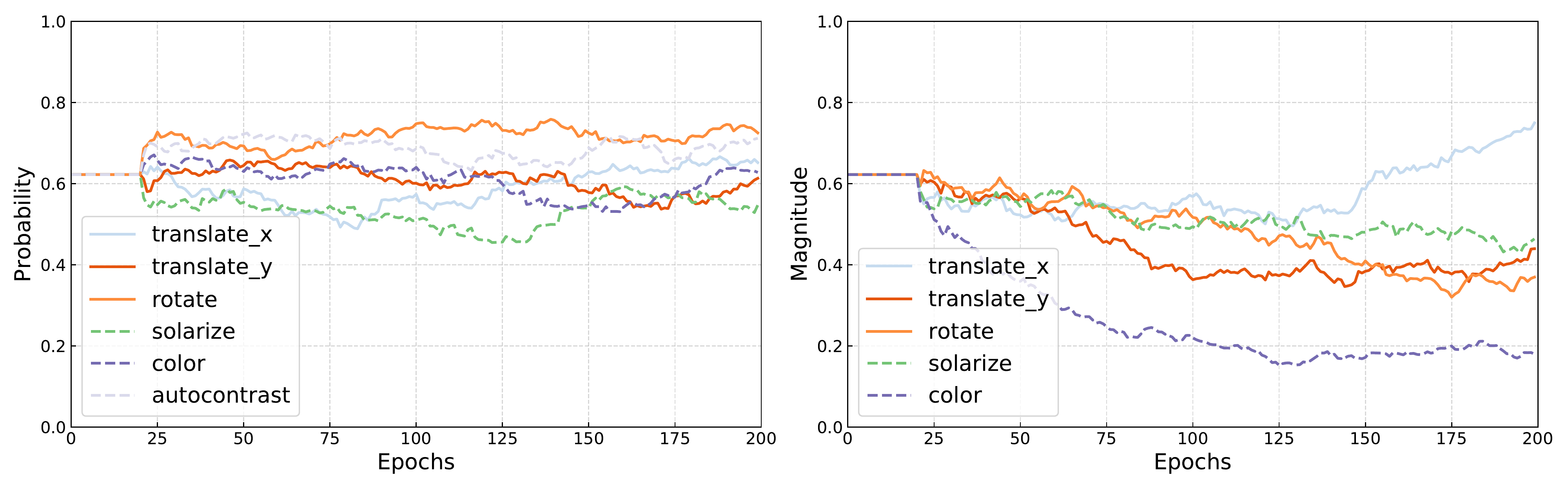}
    \caption{The development of probability parameters $p$ and magnitude parameters $\mu$ for often selected operations at the first augmentation stage (corresponds to \Cref{ap:fig:policy_development} top left) during training on CIFAR-10 with WideResNet-28-2 for 200 epochs. Note that we set the first 20 epochs to warm-up period that parameters are not updated.}
    \label{ap:fig:policy_development_prob_mag}
\end{figure}

We also present the development of probability parameters $p$ and magnitude parameters $\mu$ in \Cref{ap:fig:policy_development_prob_mag}. Interestingly, magnitude parameters diverge as training proceeds, while probability parameters remain in the range around the initial value. This observation partially agrees with the way of RandAugment, where RandAumgnet removes the probability parameter from its hyperparameters. At the same time, this results imply that the optimal magnitudes might be non-global, which disagrees with RandAugment.

\subsection{How Warm-up Affects the Performance}

\begin{figure}[t]
    \centering
    \includegraphics[width=0.5\linewidth]{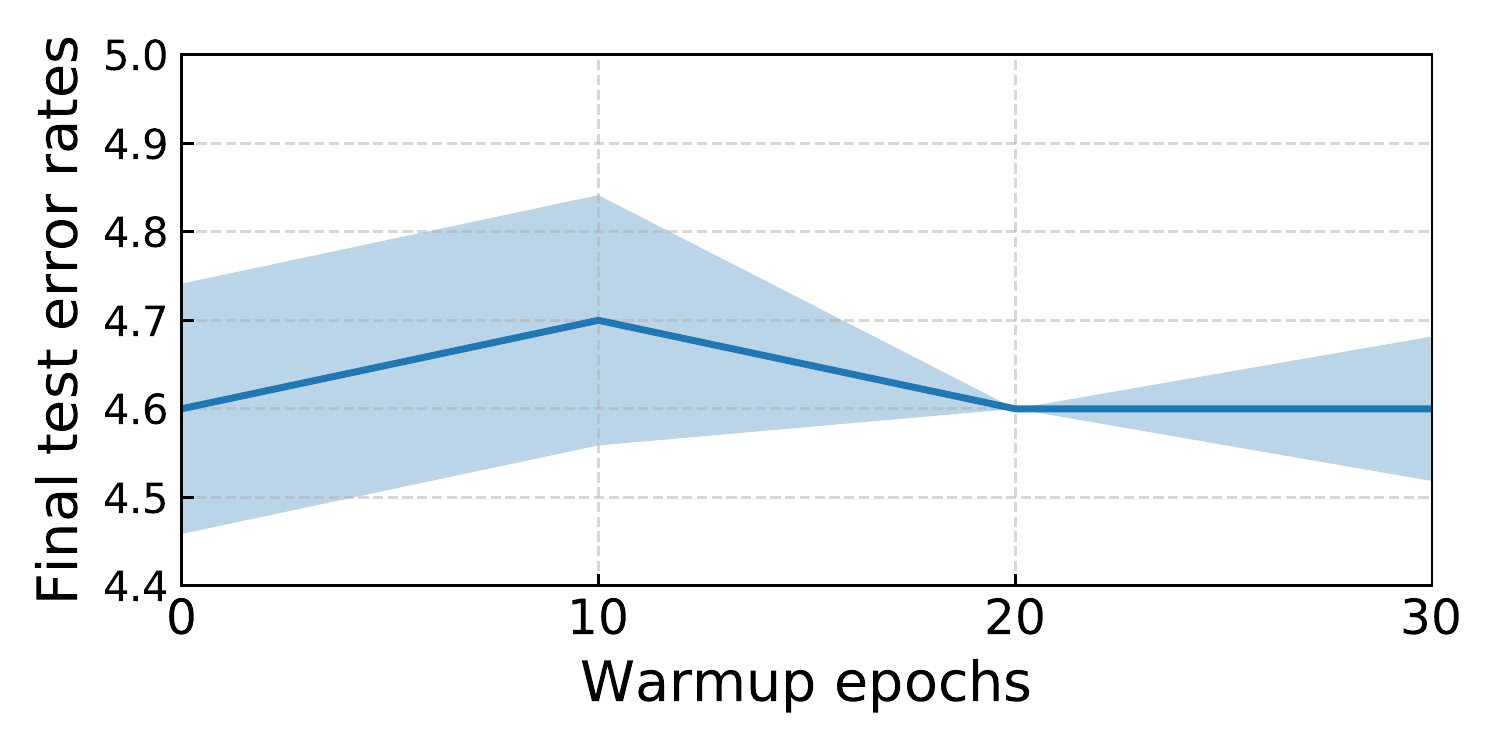}
    \caption{The relationship between the warm-up epochs and the final test error rates on CIFAR-10 with WIdeResNet-28-2. We present mean and standard deviation over three runs.}
    \label{ap:fig:warmup}
\end{figure}

\Cref{ap:fig:warmup} shows the relationship between the warm-up epochs and the final test error rates. There is no significant difference between the selection of warm-up epochs between 0 and 30.

\subsection{Comparison to Policies without Training}

 On CIFAR-10 with WideResNet-28-2, the initialized policies yield test error rates of 4.7, which is even with \randa, as intended. Therefore, policy training yields 0.1 \% of performance gain.

\end{document}

%% file: mysnipets.tex
\usepackage{amsmath,amsfonts,bm,amssymb,xspace,bbm,mathtools}

\def\eqref#1{equation~\ref{#1}}
\def\1{\mathbbm{1}}

\def\vtheta{{\bm{\theta}}}

\def\vh{{\bm{h}}}

\def\vu{{\bm{u}}}
\def\vv{{\bm{v}}}

\def\vx{{\bm{x}}}

\def\vtheta{{\bm \theta}}

\def\vpi{{\bm \pi}}

\def\vphi{{\bm \phi}}

\def\mA{{\bm{A}}}

\def\mI{{\bm{I}}}

\DeclareMathAlphabet{\mathsfit}{\encodingdefault}{\sfdefault}{m}{sl}
\SetMathAlphabet{\mathsfit}{bold}{\encodingdefault}{\sfdefault}{bx}{n}

\def\gD{\mathcal{D}}

\def\gO{\mathcal{O}}

\def\gS{\mathcal{S}}

\def\gX{\mathcal{X}}

\newcommand{\E}{\mathbb{E}}
\newcommand{\R}{\mathbb{R}}

\DeclareMathOperator*{\argmax}{argmax}
\DeclareMathOperator*{\argmin}{argmin}

\newcommand{\wrt}{w.r.t.\xspace}